\documentclass[11pt,a4paper]{article}
\usepackage[hyperref]{acl2020}
\usepackage[utf8x]{inputenc}
\usepackage{times}
\usepackage{latexsym}
\usepackage{graphicx}

\usepackage{microtype}

\usepackage{graphicx}
\usepackage{subfig}
\usepackage[T1]{fontenc}
\usepackage{epsfig}
\usepackage{caption}
\usepackage{float}
\usepackage{multirow}
\usepackage{hhline}

\usepackage[framemethod=tikz]{mdframed} 
  {\begin{mdframed}[backgroundcolor=lightgray]}%
  {\end{mdframed}}

\aclfinalcopy 

\title{How Hateful are Movies? A Study and Prediction on Movie Subtitles}

\author{Niklas von Boguszewski \thanks{ ~~~ Equal contribution} \\
 {{\small nvboguszewski@googlemail.com}} \\\And
  Sana Moin \footnotemark[1]\\
  {{\small moinsana77@gmail.com}} \\\And
  Anirban Bhowmick \footnotemark[1]\\
  { {\small anirbanbhowmick88@gmail.com}} \\ \AND
Seid Muhie Yimam \\
  { {\small seid.muhie.yimam@uni-hamburg.de}} \\ \And
  Chris Biemann \\
  { {\small christian.biemann@uni-hamburg.de}} \\\AND
   Language Technology Group\\ 
    Universit{\"a}t Hamburg, Germany\\\\}

\date{}

\begin{document}
\maketitle
\begin{abstract}
In this research, we investigate techniques to detect hate speech in movies. We introduce a new dataset collected from the subtitles of six movies, where each utterance is annotated either as hate, offensive or normal. We apply transfer learning techniques of domain adaptation and fine-tuning on existing social media datasets, namely from Twitter and Fox News. We evaluate different representations, i.e., Bag of Words (BoW), Bi-directional Long short-term memory (Bi-LSTM), and Bidirectional Encoder Representations from Transformers (BERT) on 11k movie subtitles. The BERT model obtained the best macro-averaged F1-score of 77\%. Hence, we show that transfer learning from the social media domain is efficacious in classifying hate and offensive speech in movies through subtitles.

\textbf{Cautionary Note}: The paper contains examples that many will find offensive or hateful; however, this cannot be avoided owing to the nature of the work.
\end{abstract}

\section{Introduction}
Nowadays, hate speech is becoming a pressing issue and occurs in multiple domains, mostly in the major social media platforms or political speeches. Hate speech is defined as verbal communication that denigrates a person or a community on some characteristics such as race, color, ethnicity, gender, sexual orientation, nationality, or religion \citep{Nockleby2000, davidson2017automated}. Some examples given by \citet{schmidt-wiegand-2017-survey} are:

\begin{itemize}
\item Go fucking kill yourself and die already a useless ugly pile of shit scumbag.
\item The Jew Faggot Behind The Financial Collapse.
 \item Hope one of those bitches falls over and breaks her leg.
\end{itemize}

Several sensitive comments on social media platforms have led to crime against minorities \citep{Williams2020}. Hate speech can be considered as an umbrella term that different authors have coined with different names. \citet{xu-etal-2012-learning, Hosseinmardi2015, Zhong} referred it by the term \emph{cyberbully-ing}, while \citet{davidson2017automated} used the term \emph{offensive language} to some expressions that can be strongly impolite, rude or use of vulgar words towards an individual or group that can even ignite fights or be hurtful. Use of words like f**k, n*gga, b*tch is common in social media comments, song lyrics, etc. Although these terms can be treated as obscene and inappropriate, some people also use them in non-hateful ways in different contexts \citep{davidson2017automated}. This makes it challenging for all hate speech systems to distinguish between hate speech and offensive content. \citet{davidson2017automated} tried to distinguish between the two classes in their Twitter dataset. 

These days due to globalization and online media streaming services, we are exposed to different cultures across the world through movies. Thus, an analysis of the amount of hate and offensive content in the media that we consume daily could be helpful.

Two research questions guided our research:
\begin{enumerate}
\item \label{RQclass} \textbf{RQ \ref{RQclass}}. What are the limitations of social media hate speech detection models to detect hate speech in movie subtitles? 
\item \label{RQbuild}\textbf{RQ \ref{RQbuild}}. How to build a hate and offensive speech classification model for movie subtitles?
\end{enumerate}
To address the problem of hate speech detection in movies, we chose three different models. We have used the BERT \citep{devlin-etal-2019-bert} model, due to the recent success in other NLP-related fields, a Bi-LSTM \citep{Hochreiter1997} model to utilize the sequential nature of movie subtitles and a classic Bag of Words (BoW) model as a baseline system.

The paper is structured as follows: Section \ref{relatedwork} gives an overview of the related work in this topic and Section \ref{methodology} describes the research methodology and the annotation work, while in Section \ref{dataset} we discuss the employed datasets and the pre-processing steps. 
Furthermore, Section \ref{experimentalsetup} describes the implemented models while Section \ref{discussion} presents the evaluation of the models, the qualitative analysis of the results and the annotation analysis followed by Section \ref{threat}, which covers the threats to the validity of our research. Finally, we end with the conclusion in Section \ref{conclusion} and propose further work directions in Section \ref{futurework}. 

\section{Related Work}
\label{relatedwork}
Some of the existing hate speech detection models classify comments targeted towards certain commonly attacked communities like gay, black, and Muslim, whereas in actuality, some comments did not have the intention of targeting a community \citep{Borkan2019, Dixon2018}. \citet{Mathew2021} introduced a benchmark dataset consisting of hate speech generated from two social media platforms, Twitter and Gab. In the social media space, a key challenge is to separately identify hate speech from offensive text. Although they might appear the same way semantically, they have subtle differences. Therefore they tried to solve the bias and interpretability aspect of hate speech and did a three-class classification (i.e., hate, offensive, or normal). They reported the best macro-averaged F1-score of 68.7\% on their BERT-HateXplain model. It is also one of the models that we use in our study, as it is one of the `off-the-shelf` hate speech detection models that can easily be employed for the topic at hand.

Lexicon-based detection methods have low precision because they classify the messages based on the presence of particular hate speech-related terms, particularly those insulting, cursing, and bullying words. \citet{davidson2017automated} used a crowdsourced hate speech lexicon to identify tweets with the occurrence of hate speech keywords to filter tweets. They then used crowdsourcing to label these tweets into three classes: hate speech, offensive language, and neither. In their dataset, the more generic racist and homophobic tweets were classified as hate speech, whereas the ones involving sexist and abusive words were classified as offensive. It is one of the datasets we have used in exploring transfer learning and model fine-tuning in our study.
 
Due to global events, hate speech also plagues online news platforms. In the news domain, context knowledge is required to identify hate speech. \citet{gao2017} conducted a study on a dataset prepared from user comments on news articles from the Fox News platform. It is the second dataset we have used to explore transfer learning from the news domain to movie subtitles in our study.

Several other authors have collected the data from different online platforms and labeled them manually. Some of these data sources are: Twitter \citep{Xiang2012, xu-etal-2012-learning}, Instagram \citep{Hosseinmardi2015, Zhong}, Yahoo! \citep{Nobata2016, Djuric2015}, YouTube \citep{Dinakar2012} and Whisper \citep{silva2016analyzing} to name a few. Most of the data sources used in the previous studies are based on social media, news, and micro-blogging platforms. However, the notion of the existence of hate speech in movie dialogues has been overlooked. Thus in our study, we first explore how the different existing ML (Machine Learning) models classify hate and offensive speech in movie subtitles and propose a new dataset compiled from six movie subtitles. 

\section{Research Methodology}
\label{methodology}
To investigate the problem of detecting hate and offensive speech in movies, we used different machine learning models trained on social media content such as tweets or discussion thread comments from news articles. Here, the models in our research were developed and evaluated on an in-domain 80\% train and 20\% test split data using the same random state to ensure comparability. 

We have developed six different models: two Bi-LSTM models, two BoW models, and two BERT models. For each pair, one of them has been trained on a dataset consisting of Twitter posts and the other on a dataset consisting of Fox News discussion threads. The trained models have been used to classify movie subtitles to evaluate their performance by domain adaptation from social media content to movies. In addition, another state-of-the-art BERT-based classification model called \textit{HateXplain} \cite{Mathew2021} has been used to classify the movies out of the box. While it is also possible to further fine-tune the HateXplain model, we are restricted in reporting the result of the 'off-the-shelf' classification system to new domains, such as movie subtitles.

Furthermore, the movie dataset we have collected (see Section \ref{Datasets}) is used to train domain-specific BoW, Bi-LSTM, and BERT models using 6-fold cross-validation, where each movie was selected as a fold and report the averaged results. 
Finally, we have identified the best model trained on social media content based on macro-averaged F1-score and fine-tuned it with the movie dataset using 6-fold cross-validation on that particular model, to investigate fine-tuning and transfer learning capabilities for hate speech on movie subtitles.

\subsection{Annotation Guidelines}

In our annotation guidelines, we defined hateful speech as a language used to express hatred towards a targeted individual or group or is intended to be derogatory, to humiliate, or to insult the members of the group, based on attributes such as race, religion, ethnic origin, sexual orientation, disability, or gender. Although the meaning of hate speech is based on the context, we provided the above definition agreeing to the definition provided by \citet{Nockleby2000, davidson2017automated}. Offensive speech uses profanity, strongly impolite, rude, or vulgar language expressed with fighting or hurtful words to insult a targeted individual or group \citep{davidson2017automated}. We used the same definition also for offensive speech in the guidelines. The remaining subtitles were defined as normal.  

\subsection{Annotation Task}
For the annotation of movie subtitles, we have used Amazon Mechanical Turk (MTurk) crowdsourcing. Before the main annotation task, we have conducted an annotation pilot study, where 40 subtitles texts were randomly chosen from the movie subtitle dataset. Each of them has included 10 hate speech, 10 offensive, and 20 normal subtitles that are manually annotated by experts. In total, 100 MTurk workers were assigned for the annotation task. We have used the built-in MTurk qualification requirement (HIT approval rate higher than 95\% and number of HITs approved larger than 5000) to recruit workers during the Pilot task. Each worker was assessed for accuracy and the 13 workers who have completed the task with the highest annotation accuracy were chosen for the main study task. The rest of the workers were compensated for the task they have completed in the pilot study and blocked from participating in the main annotation task. For each HIT, the workers are paid 40 cents both for the pilot and the main annotation task.

\begin{figure}[H]
   {\includegraphics[width=0.45\textwidth]{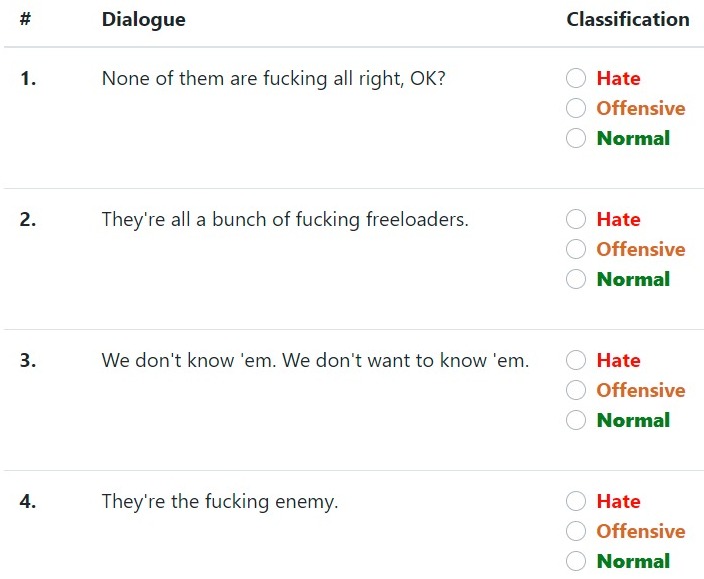}}
   \caption{Annotation template containing a batch of the movie American History X}\label{AnnotationTempalte}
\end{figure}

For the main task, the 13 chosen MTurk workers were first assigned to one movie subtitle annotation to further look at the annotator agreement as will be described in Section \ref{AnnotationAnalysis}. Two annotators were replaced during the main annotation task with the next-best workers from the identified workers in the pilot study. This process was repeated after each movie annotation for the remaining five movies. One batch consists of 40 subtitles which were displayed in chronological order to the worker. Each batch has been annotated by three workers. In Figure \ref{AnnotationTempalte}, you can see the first four questions of a batch out of the movie American History X \citeyear{AmericanHistoryX}. 

\graphicspath{{images/}}

\section{Datasets}\label{Datasets}
\label{dataset}
\begin{table}
\centering
\begin{tabular}{lrrrr}
\hline \textbf{Dataset} & \textbf{normal} & \textbf{offensive} & \textbf{hate} & \textbf{\# total}\\ \hline
Twitter & 0.17 & 0.78 & 0.05  & 24472 \\
Fox News & 0.72 & - & 0.28 & 1513\\
Movies & 0.84 & 0.13 & 0.03  & 10688\\

\hline
\end{tabular}
\caption{\label{TotalDataset} Class distribution for the different datasets}
\end{table}

\begin{table}
\centering
\resizebox{0.5\textwidth}{!}{%
\begin{tabular}{lrrrr}
\hline \textbf{Name} & \textbf{normal} & \textbf{offensive} & \textbf{hate} & \textbf{\#total} \\ \hline
Django Unchained & 0.89 & 0.05 & 0.07 & 1747 \\
BlacKkKlansman & 0.89 & 0.06 & 0.05 & 1645 \\
American History X  & 0.83 & 0.13 & 0.03 & 1565\\
Pulp Fiction & 0.82 & 0.16 & 0.01 & 1622\\
South Park & 0.85 & 0.14 & 0.01 & 1046 \\
The Wolf of Wall Street & 0.81 & 0.19 & 0.001 & 3063 \\

\hline
\end{tabular}
}
\caption{\label{MovieDataset} Class distribution on each movie}
\end{table}
The publicly available Fox News corpus\footnote{\url{https://github.com/sjtuprog/fox-news-comments}} consists of 1,528 annotated comments compiled from ten discussion threads that happened on the Fox News website in 2016. The corpus does not differentiate between offensive and hateful comments. This corpus has been introduced by \citet{gao2017} and has been annotated by two trained native English speakers. We have identified 13 duplicates and two empty comments in this corpus and removed them for accurate training results. The second publicly available corpus we use consists of 24,802 tweets\footnote{\url{https://github.com/t-davidson/hate-speech-and-offensive-language/}}. We identified 204 of them as duplicates and removed them again to achieve accurate training results. The corpus has been introduced by \citet{davidson2017automated} and was labeled by CrowdFlower workers as hate speech, offensive, and neither. The last class is referred to as \textit{normal} in this paper. The distribution of the normal, offensive, and hate classes can be found in Table \ref{TotalDataset}.

The novel movie dataset we introduce consists of six movies. The movies have been chosen based on keyword tags provided by the IMDB website\footnote{\url{https://www.imdb.com/search/keyword/}}. The tags \textit{hate-speech} and \textit{racism} were chosen because we assumed that they were likely to contain a lot of hate and offensive speech. The tag \textit{friendship} was chosen to get contrary movies containing a lot of normal subtitles, with less hate speech content. In addition, we excluded movie genres like documentations, fantasy, or musicals to keep the movies comparable to each other. Namely we have chosen the movies BlacKkKlansman \citeyearpar{BlacKkKlansman} which was tagged as \textit{hate-speech}, Django Unchained \citeyearpar{DjangoUnchained}, American History X \citeyearpar{AmericanHistoryX} and Pulp Fiction \citeyearpar{PulpFiction} which were tagged as \textit{racism} whereas South Park \citeyearpar{SouthPark} as well as The Wolf of Wall Street \citeyearpar{WolfOfWallStreet} were tagged as \textit{friendship} in December 2020. The detailed distribution of the normal, offensive, and hate classes, movie-wise, can be found in Table \ref{MovieDataset}.

\subsection{Pre-processing}
\label{preprossessing}
The goal of the pre-processing step was to make the text of the Tweets and conversational discussions as comparable as possible to the movie subtitles since we assume that this will improve the transfer learning results. Therefore, we did not use pre-processing techniques like stop word removal or lemmatization.

\subsection{Data Cleansing}
After performing a manual inspection, we applied certain rules to remove the textual noise from our datasets. The following was the noise observed in each dataset, which we removed for the Twitter and Fox News datasets: (1) repeated punctuation marks, (2) multiple username tags, (3) emoticon character encodings, and (4) website links.
For the movie subtitle text dataset: (1) sound expressions, e.g [PEOPLE CHATTERING], [DOOR OPENING],
    (2) name header of the current speaker, e.g. "DIANA: Hey, what's up?" which refers to Diana is about to say something,
    (3) HTML tags,
   (4) non-alpha character subtitle, and
    (5) non-ASCII characters.

\subsection{Subtitle format conversion}

The downloaded subtitle files are provided by the website \textit{www.opensubtitles.org}\footnote{\url{https://www.opensubtitles.org/}} and are free to use for scientific purposes. The files are available in the SRT-format\footnote{\url{https://en.wikipedia.org/wiki/SubRip}} that have a time duration along with a subtitle, which while watching appears on the screen in a given time frame. We performed the following operations to create the movie dataset: (1) Converted the SRT-format to CSV-format by separating start time, end time, and the subtitle text, (2) Fragmented subtitles which were originally single appearances on the screen and spanned across multiple screen frames were combined, by identifying sentence-ending punctuation marks, 
    (3) Combined single word subtitles with the previous subtitle because single word subtitles tend to be expressions to what has been said before.

\section{Experimental Setup}
\label{experimentalsetup}
The Bi-LSTM models are built using the Keras and the BoW models are built using the PyTorch library while both are trained with a 1e-03 learning rate and categorical cross-entropy loss function. 

For the development of BERT-based models, we rely on the \textit{TFBERTForSequenceClassification} algorithm, which is provided by HuggingFace\footnote{\url{https://huggingface.co/transformers}} and pre-trained on \textit{bert-base-uncased}. Learning rate of 3e-06 and sparse categorical cross-entropy loss function was used for this. 
All the models used the Adam optimizer \cite{kingma2017adam}.
We describe the detailed hyper-parameters for all the models used for all the experiments in the Appendix \ref{section:hyperparams}.

\section{Results and Annotation Analysis} 
\label{discussion}

In this section, we will discuss the different classification results obtained from the various hate speech classification models. We will also briefly present a qualitative exploration of the annotated movie datasets. The model referred in the tables as LSTM refers to Bi-LSTM models used. 
\subsection{Classification results and Discussion}
We have introduced a new dataset of movie subtitles in the field of hate speech research. A total of six movies are annotated, which consists of sequential subtitles.

First, we experimented on the \textit{HateXplain} model \cite{Mathew2021} by testing the model's performance on the movie dataset. We achieved a macro-averaged F1-score of 66\% (see Table \ref{hateXplain}). Next, we tried to observe how the different models (BoW, Bi-LSTM, and BERT) perform using transfer learning and how comparable are those results to this state-of-the-art model's results. 

\begin{table}
\centering

\begin{tabular}{|p{2.2cm}|p{1.6cm}|p{1cm}|p{1cm}|}
\hline
\textbf{Model} & \textbf{Class} & \textbf{F1-Score} & \textbf{Macro AVG F1}\\
\hhline{|=|=|=|=|}
HateXplain & normal & 0.93 & \multirow{3}{\columnwidth}{0.66} \\
HateXplain & offensive & 0.27 &  \\
HateXplain & hate & 0.77 & \\
\hline
\end{tabular}

\caption{\label{hateXplain}
Prediction results using the HateXplain model on the movie dataset (domain adaptation)
}
\end{table}

We trained and tested the BERT, Bi-LSTM, and BoW model by applying an 80:20 split on the social media datasets (see Table \ref{8020split}). 
When applied to the Fox News dataset, we observed that BERT performed better than both BoW and Bi-LSTM with a small margin in terms of macro-averaged F1-score. Hate is detected close to 50\% whereas normal is detected close to 80\% for all three models on F1-score.

When applied on the Twitter dataset, results are almost the same for the BoW and Bi-LSTM models, whereas the BERT model performed close to 10\% better by reaching a macro-averaged F1-score of 76\%. 
All the models have a high F1-score of above 90\% for identifying offensive class. This goes along with the fact that the offensive class is the dominant one in the Twitter dataset (Table \ref{TotalDataset}).

Hence, by looking at the macro-averaged F1-score values, BERT performed best in the task for training and testing on social media content on both datasets.


\begin{table}
\centering
\begin{tabular}{|p{1.6cm}|p{1cm}|p{1.4cm}|p{0.8cm}|p{0.95cm}|}
\hline
\textbf{Dataset} & \textbf{Model} & \textbf{Class} & \textbf{F1-Score} & \textbf{Macro AVG F1}\\
\hhline{|=|=|=|=|=|}
\multirow{6}{*}{Fox News} & \multirow{2}{*}{BoW} & normal & 0.83 & \multirow{2}{\columnwidth}{0.63}\\
&   & hate  & 0.43 & \\\cline{2-5}
& \multirow{2}{*}{BERT} & normal  & 0.86 & \multirow{2}{\columnwidth}{\textbf{0.68}}\\
&  & hate  & 0.51 & \\\cline{2-5}
& \multirow{2}{*}{LSTM} & normal  & 0.77 & \multirow{2}{\columnwidth}{0.62}\\
&  & hate  & 0.46 & \\\cline{2-5}
\hline
\hhline{|=|=|=|=|=|}
\multirow{9}{*}{Twitter} & \multirow{3}{*}{BoW} & normal  & 0.78 & \multirow{3}{\columnwidth}{0.66}\\
 &  & offensive  & 0.93 &\\
 &  & hate  & 0.26 & \\\cline{2-5}

 & \multirow{3}{*}{BERT} & normal & 0.89 & \multirow{3}{\columnwidth}{\textbf{0.76}}\\
 &  & offensive  & 0.95 & \\
 &  & hate  & 0.43 & \\\cline{2-5}

 &  \multirow{3}{*}{LSTM} & normal & 0.76 & \multirow{3}{\columnwidth}{0.66}\\
 &  & offensive  & 0.91 & \\
 &  & hate  & 0.31 & \\\cline{2-5}
\hline
\end{tabular}
\caption{\label{8020split}
In-domain results on Twitter and Fox News with 80:20 split}
\end{table}

Next, we train on social media data and test on the six movies (see Table \ref{DomainChange}) to address RQ \ref{RQclass}.

When trained on the Fox News dataset, BoW and Bi-LSTM performed similarly by poorly detecting hate in the movies. In contrast, BERT identified the hate class more than twice as well by reaching an F1-score of 39\%. 

When trained on the Twitter dataset, BERT performed almost double in terms of macro-averaged F1-score than the other two models. 
Even though the detection for the offensive class was high on the Twitter dataset (see Table \ref{8020split}) the models did not perform as well on the six movies, which could be due to the domain change. However, BERT was able to perform better on the hate class, even though it was trained on a small proportion of hate content in the Twitter dataset. The other two models performed very poorly.




\begin{table}
\centering

\begin{tabular}{|p{1.6cm}|p{1cm}|p{1.4cm}|p{0.8cm}|p{0.95cm}|}
\hline
\textbf{Dataset} & \textbf{Model} & \textbf{Class} & \textbf{F1-Score} & \textbf{Macro AVG F1}\\
\hhline{|=|=|=|=|=|}
\multirow{6}{*}{Fox News} & \multirow{2}{*}{BoW} & normal & 0.86 & \multirow{2}{\columnwidth}{0.51}\\
 &  & hate  & 0.15 & \\\cline{2-5}
 
 & \multirow{2}{*}{BERT} & normal & 0.89 & \multirow{2}{\columnwidth}{\textbf{0.64}}\\
 &  & hate & 0.39 & \\\cline{2-5}

 & \multirow{2}{*}{LSTM} & normal & 0.83 & \multirow{2}{\columnwidth}{0.51}\\
 &  & hate & 0.18 &\\\cline{2-5}

\hhline{|=|=|=|=|=|}
\multirow{9}{*}{Twitter} & \multirow{3}{*}{BoW} & normal & 0.62 & \multirow{3}{\columnwidth}{0.37}\\
 &  & offensive & 0.32 & \\
 &  & hate & 0.15 & \\\cline{2-5}

 & \multirow{3}{*}{BERT}  & normal & 0.95 & \multirow{3}{\columnwidth}{\textbf{0.77}}\\
 &  & offensive & 0.74 &\\
 &  & hate & 0.63 & \\\cline{2-5}

 & \multirow{3}{*}{LSTM}  & normal & 0.66 & \multirow{3}{\columnwidth}{0.38}\\
 &  & offensive & 0.34 & \\
 &  & hate & 0.16 & \\\cline{2-5}
\hline
\end{tabular}

\caption{\label{DomainChange}
Prediction results using the models trained on social media content to classify the six movies (domain adaptation)
}
\end{table}

To address RQ \ref{RQbuild}, we train new models from scratch on the six movies dataset using 6-fold cross-validation (see Table \ref{CrossValidation}). In this setup, each fold represents one movie that is exchanged iteratively during evaluation. 

Compared to the domain adaptation (see Table \ref{DomainChange}), the BoW and Bi-LSTM models performed better. 
Bi-LSTM distinguished better than BoW among hate and offensive while maintaining a good identification of the normal class resulting in a better macro-averaged F1-score of 71\% as compared to 64\% for the BoW model.
BERT performed best across all three classes resulting in 10\% better results compared to the Bi-LSTM model on macro-averaged F1-score, however, it has similar results when compared to the domain adaptation (see Table \ref{DomainChange}) results.

Furthermore, the absolute amount of hateful subtitles in the movies The Wolf of Wall Street (3), South Park (10), and Pulp Fiction (16) are very minor, hence the cross-validation on these three movies as test set is very sensible of only predicting a few of them wrong since a few of them will already result in a high relative amount.

\begin{table}
\centering
\begin{tabular}{|p{1.2cm}|p{1.4cm}|p{1.4cm}|p{0.8cm}|p{0.95cm}|}
\hline
\textbf{Dataset} & \textbf{Model} & \textbf{Class} & \textbf{F1-Score} & \textbf{Macro AVG F1}\\
\hhline{|=|=|=|=|=|}
\multirow{9}{*}{Movies} & \multirow{3}{*}{BoW} & normal & 0.95 & \multirow{3}{\columnwidth}{0.64}\\
 &  & offensive & 0.59 &\\
 &  & hate & 0.37 & \\\cline{2-5}

 & \multirow{3}{*}{BERT} & normal & 0.97 & \multirow{3}{\columnwidth}{\textbf{0.81}}\\
 &  & offensive & 0.76 & \\
 &  & hate & 0.68 & \\\cline{2-5}

 & \multirow{3}{*}{LSTM} & normal & 0.95 & \multirow{3}{\columnwidth}{0.71}\\
 &  & offensive & 0.63 & \\
 &  & hate & 0.56 & \\\cline{2-5}
\hline
\end{tabular}
\caption{\label{CrossValidation}
In-domain results using models trained on the movie dataset using 6-fold cross-validation
}
\end{table}

We have also tried to improve our BERT model trained on social media content (Table \ref{8020split}) by fine-tuning it via 6-fold cross-validation using the six movies dataset (see Table \ref{crossvalfinetuned}).

The macro-averaged F1-score increased compared to the domain adaptation (see Table \ref{DomainChange}) from 64\% to 89\% for the model trained on the Fox News dataset.
For the Twitter dataset the macro-averaged F1-score is comparable to the domain adaptation (see Table \ref{DomainChange}) and in-domain results (see Table \ref{CrossValidation}). 
Compared to the results of the HateXplain model (see Table \ref{hateXplain}) the identification of the normal utterances are comparable whereas the offensive class was identified by our BERT model much better, with an increment of 48\%, but the hate class was identified by a decrement of 18\%.

The detailed results of all experiments is given in Appendix  \ref{section:detailmetrics}.

\subsection{Qualitative Analysis}
\begin{figure}[H]
   {\includegraphics[width=0.45\textwidth]{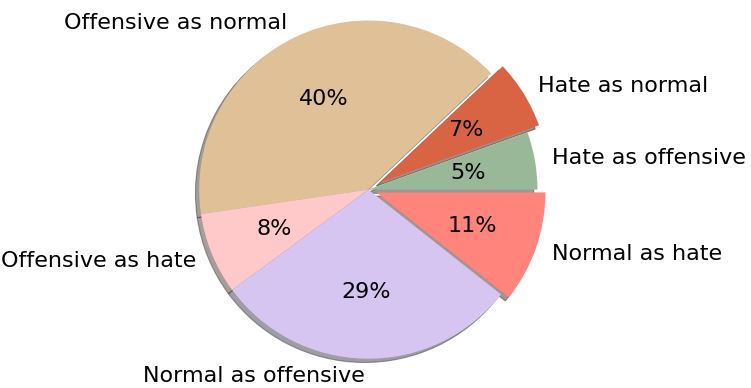}}
   \caption{Label misclassification on the movie dataset using the BERT model of Table \ref{crossvalfinetuned}\label{labelMisclass} trained on the Twitter dataset}
\end{figure}

In this section, we investigate the unsuccessfully classified utterances (see Figure \ref{labelMisclass}) of all six movies by the BERT model trained on the Twitter dataset and fine-tuned with the six movies via 6-fold cross-validation (see Table \ref{crossvalfinetuned}) to analyze the model addressing RQ \ref{RQbuild}.

The majority of unsuccessfully classified utterances (564) are offensive classified as normal and vice versa resulting in 69\%. Hate got classified as offensive in 5\% of all cases and offensive as hate in 8\%. The remaining misclassification is between normal and hate resulting in 18\%, which we refer to as the most critical for us to analyze further. 

We looked at the individual utterances of the hate class misclassified as normal (37 utterances). We observed that most of them were sarcastic and those did not contain any hate keywords, whereas some could have been indirect or context-dependent, for example, the utterance \textit{"It's just so beautiful. We're cleansing this country of a backwards race of chimpanzees"} indirectly and sarcastically depicts hate speech which our model could not identify. We assume that our model has shortcomings in interpreting those kinds of utterances correctly.

Furthermore, we analyzed the utterances of the class normal which were misclassified as hate (60 utterances). We observed that around a third of them were actual hate but were misclassified by our annotators as normal, hence those were correctly classified as hate by our model. We noticed that a fifth of them contain the keyword \textit{"Black Power"}, which we refer to as normal whereas the BERT model classified them as hate.


%

\begin{table}
\centering
\begin{tabular}{|p{1.1cm}|p{1.65cm}|p{1.25cm}|p{0.8cm}|p{0.95cm}|}
\hline
\textbf{Dataset} & \textbf{Model} & \textbf{Class} & \textbf{F1-Score} & \textbf{Macro AVG F1}\\
\hhline{|=|=|=|=|=|}
\multirow{5}{*}{Movies} & \multirow{1}{*}{BERT} & normal & 0.97 & \multirow{2}{\columnwidth}{0.89}\\
 & (Fox News) & hate & 0.82 & \\\cline{2-5}
 
 & \multirow{2}{*}{BERT} & normal & 0.97 & \multirow{3}{\columnwidth}{\textbf{0.77}}\\
 &  & offensive & 0.75 & \\
 & (Twitter) & hate & 0.59 & \\
\hline
\end{tabular}
\caption{\label{crossvalfinetuned}
Prediction results using BERT models trained on the Twitter and Fox News datasets and fine-tuned them with the movie dataset by applying 6-fold cross-validation (fine-tuning)
}
\end{table}

\subsection{Annotation Analysis} \label{AnnotationAnalysis}

Using the MTurk crowdsourcing, a total of 10,688 subtitles (from the six movies) are annotated. For each of the three workers involved, 81\% agreed to the same class. Out of the total annotations, only 0.7\% received disagreement on the classes (where all the three workers chose a different class for each subtitle).

To ensure the quality of the classes for the training, we chose majority voting. In the case of disagreement, we took the offensive class as the final class of the subtitle. 
One reason why workers do disagree might be that they do interpret a scene differently. We think that providing the video and audio clips of the subtitle frames might help to disambiguate such confusions.

Let us consider an example from one of the annotation batches that describes a scene where the shooting of an Afro-American appears to happen. Subtitle 5 in that batch reads out \textit{"Shoot the nigger!"}, and subtitle 31 states \textit{"Just shit. Got totally out of control."}, which was interpreted as normal by a worker who might not be sensible to the word \textit{shit}, as offensive speech by a worker who is, in fact, sensible to the word \textit{shit} or as hate speech by a worker who thinks that the word \textit{shit} refers to the Afro-American.

The movie Django Unchained \citeyear{DjangoUnchained} was tagged as \textit{racism} and has been annotated as the most hateful movie (see Table \ref{MovieDataset}) followed by BlacKkKlansman \citeyear{BlacKkKlansman} and American History X \citeyear{AmericanHistoryX} which where tagged as \textit{racism} or \textit{hateful}. This indicates that hate speech and racist comments often go along together. 
As expected, movies tagged by \textit{friendship} like The Wolf of Wall Street \citeyear{WolfOfWallStreet} and South Park \citeyear{SouthPark} were less hateful. 
Surprisingly the percentage of offensive speech increases when the percentage of hate decreases making the movies tagged by \textit{friendship} most offensive in our movie dataset. 

\section{Threats to Validity}
\label{threat}
\begin{enumerate}
\item The pre-processing of the movies or the social media datasets could have deleted crucial parts which would have made a hateful tweet normal, for example. Thus the training on such datasets could impact the training negatively.
\item Movies are not real, they are more like a very good simulation. Thus, for this matter, hate speech is simulated and arranged. 
Maybe documentation movies are better suited since they tend to cover real-case scenarios. 
\item The annotations could be wrong since the task of identifying hate speech is subjective. 
\item Movies might not contain a lot of hate speech, hence the need to detect them is very minor.
\item As the annotation process was done batch-wise, annotators might lose crucial contextual information when the batch change happens, as it misses the chronological order of the dialogue. 
\item Only textual data might not provide enough contextual information for the annotators to correctly annotate the dialogues as the other modalities of the movies (audio and video) are not considered.
\end{enumerate}

\section{Conclusion}
\label{conclusion}

In this paper, we applied different approaches to detect hate and offensive speech in a novel proposed movie subtitle dataset. In addition, we proposed a technique to combine fragments of movie subtitles and made the social media text content more comparable to movie subtitles (for training purposes). 

For the classification, we used two techniques of transfer learning, i.e., domain adaptation and fine-tuning. The former was used to evaluate three different ML models, namely Bag of Words for a baseline system, transformer-based systems as they are becoming the state-of-the-art classification approaches for different NLP tasks, and Bi-LSTM-based models as our movie dataset represents sequential data for each movie. The latter was performed only on the BERT model and we report our best result by cross-validation on the movie dataset. 

All three models were able to perform well for the classification of the normal class. Whereas when it comes to the differentiation between offensive and hate classes, BERT achieved a substantially higher F1-score as compared to the other two models.

The produced artifacts could have practical significance in the field of movie recommendations. We will release the annotated datasets, keeping all the contextual information (time offsets of the subtitle, different representations, etc.), the fine-tuned and newly trained models, as well as the python source code and pre-processing scripts, to pursue research on hate speech on movie subtitles.\footnote{\url{https://github.com/uhh-lt/hatespeech}} 

\section{Further Work}
\label{futurework}
The performance of hate speech detection in movies can be improved by increasing the existing movie dataset with movies that contain a lot of hate speech.
Moreover, multi-modal models can also improve performance by using speech or image. In addition, some kind of hate speech can only be detected through the combination of different modals, like some memes in the hateful meme challenge by Facebook \citep{FacebookHateMeme2021} e.g. a picture that says \textit{look how many people love you} whereas the image shows an empty desert. \\
Furthermore, we also did encounter the widely reported sparsity of hate speech content, which can be mitigated by using techniques such as data augmentation, or balanced class distribution. We intentionally did not perform shuffling of all six movies before splitting into k-folds to retain a realistic scenario where a classifier is executed on a new movie.

Another interesting aspect that can be looked at is the identification of the target groups of the hate speech content in movies and to see the more prevalent target groups.
This work can also be extended for automated annotation of movies to investigate the distribution of offensive and hate speech.

\bibliography{acl2020}

\begin{thebibliography}{26}
\expandafter\ifx\csname natexlab\endcsname\relax\def\natexlab#1{#1}\fi

\bibitem[{Pul(1994)}]{PulpFiction}
 1994.
\newblock \href
  {https://www.opensubtitles.org/de/search/sublanguageid-eng/idmovie-60}
  {{Movie: Pulp Fiction}}.
\newblock Last visited 23.05.2021.

\bibitem[{Ame(1998)}]{AmericanHistoryX}
 1998.
\newblock \href
  {https://www.opensubtitles.org/de/search/sublanguageid-eng/idmovie-1583}
  {{Movie: American History X}}.
\newblock Last visited 23.05.2021.

\bibitem[{Sou(1999)}]{SouthPark}
 1999.
\newblock \href
  {https://www.opensubtitles.org/de/search/sublanguageid-eng/idmovie-3053}
  {{Movie: South Park: Bigger, Longer \& Uncut}}.
\newblock Last visited 23.05.2021.

\bibitem[{Dja(2012)}]{DjangoUnchained}
 2012.
\newblock \href
  {https://www.opensubtitles.org/de/search/sublanguageid-eng/idmovie-131086}
  {{Movie: Django Unchained}}.
\newblock Last visited 23.05.2021.

\bibitem[{Wol(2013)}]{WolfOfWallStreet}
 2013.
\newblock \href
  {https://www.opensubtitles.org/de/search/sublanguageid-eng/idmovie-161460}
  {{Movie: The Wolf of Wall Street}}.
\newblock Last visited 23.05.2021.

\bibitem[{Bla(2018)}]{BlacKkKlansman}
 2018.
\newblock \href
  {https://www.opensubtitles.org/de/search/sublanguageid-eng/idmovie-639472}
  {{Movie: BlacKkKlansman}}.
\newblock Last visited 23.05.2021.

\bibitem[{Borkan et~al.(2019)Borkan, Dixon, Sorensen, Thain, and
  Vasserman}]{Borkan2019}
Daniel Borkan, Lucas Dixon, Jeffrey Sorensen, Nithum Thain, and Lucy Vasserman.
  2019.
\newblock {Nuanced Metrics for Measuring Unintended Bias with Real Data for
  Text Classification}.
\newblock In \emph{Companion of The 2019 World Wide Web Conference, WWW}, pages
  491--500, San Francisco, CA, USA.

\bibitem[{Davidson et~al.(2017)Davidson, Warmsley, Macy, and
  Weber}]{davidson2017automated}
Thomas Davidson, Dana Warmsley, Michael Macy, and Ingmar Weber. 2017.
\newblock {Automated Hate Speech Detection and the Problem of Offensive
  Language}.
\newblock In \emph{Proceedings of the 11th International AAAI Conference on Web
  and Social Media}, pages 512--515, Montr\'{e}al, QC, Canada.

\bibitem[{Devlin et~al.(2019)Devlin, Chang, Lee, and
  Toutanova}]{devlin-etal-2019-bert}
Jacob Devlin, Ming-Wei Chang, Kenton Lee, and Kristina Toutanova. 2019.
\newblock \href {https://doi.org/10.18653/v1/N19-1423} {{BERT: Pre-training of
  Deep Bidirectional Transformers for Language Understanding}}.
\newblock In \emph{Proceedings of the 2019 Conference of the North {A}merican
  Chapter of the Association for Computational Linguistics: Human Language
  Technologies, Volume 1 (Long and Short Papers)}, pages 4171--4186,
  Minneapolis, MN, USA. Association for Computational Linguistics.

\bibitem[{Dinakar et~al.(2012)Dinakar, Jones, Havasi, Lieberman, and
  Picard}]{Dinakar2012}
Karthik Dinakar, Birago Jones, Catherine Havasi, Henry Lieberman, and Rosalind
  Picard. 2012.
\newblock \href {https://doi.org/10.1145/2362394.2362400} {{Common Sense
  Reasoning for Detection, Prevention, and Mitigation of Cyberbullying}}.
\newblock \emph{ACM Trans. Interact. Intell. Syst.}, 2(3):18:1--18:30.

\bibitem[{Dixon et~al.(2018)Dixon, Li, Sorensen, Thain, and
  Vasserman}]{Dixon2018}
Lucas Dixon, John Li, Jeffrey Sorensen, Nithum Thain, and Lucy Vasserman. 2018.
\newblock {Measuring and Mitigating Unintended Bias in Text Classification}.
\newblock In \emph{Proceedings of the 2018 AAAI/ACM Conference on AI, Ethics,
  and Society. Association for Computing Machinery}, pages 67–--73, New
  Orleans, LA, USA.

\bibitem[{Djuric et~al.(2015)Djuric, Zhou, Morris, Grbovic, Radosavljevic, and
  Bhamidipati}]{Djuric2015}
Nemanja Djuric, Jing Zhou, Robin Morris, Mihajlo Grbovic, Vladan Radosavljevic,
  and Narayan Bhamidipati. 2015.
\newblock {Hate Speech Detection with Comment Embeddings}.
\newblock In \emph{Proceedings of the 24th International Conference on World
  Wide Web}, pages 29--30, Florence, Italy.

\bibitem[{Hochreiter and Schmidhuber(1997)}]{Hochreiter1997}
Sepp Hochreiter and J\"{u}rgen Schmidhuber. 1997.
\newblock \href {https://doi.org/10.1162/neco.1997.9.8.1735} {{Long Short-Term
  Memory}}.
\newblock \emph{Neural Comput.}, 9(8):1735–--1780.

\bibitem[{Hosseinmardi et~al.(2015)Hosseinmardi, Mattson, Rafiq, Han, Lv, and
  Mishra}]{Hosseinmardi2015}
Homa Hosseinmardi, Sabrina~Arredondo Mattson, Rahat~Ibn Rafiq, Richard Han, Qin
  Lv, and Shivakant Mishra. 2015.
\newblock \href {http://arxiv.org/abs/1503.03909} {{Detection of Cyberbullying
  Incidents on the Instagram Social Network}}.
\newblock \emph{CoRR}, abs/1503.03909.

\bibitem[{Kiela et~al.(2020)Kiela, Firooz, Mohan, Goswami, Singh, Ringshia, and
  Testuggine}]{FacebookHateMeme2021}
Douwe Kiela, Hamed Firooz, Aravind Mohan, Vedanuj Goswami, Amanpreet Singh,
  Pratik Ringshia, and Davide Testuggine. 2020.
\newblock \href {http://arxiv.org/abs/2005.04790} {{The Hateful Memes
  Challenge: Detecting Hate Speech in Multimodal Memes}}.
\newblock \emph{CoRR}, abs/2005.04790.

\bibitem[{Kingma and Ba(2015)}]{kingma2017adam}
Diederik~P. Kingma and Jimmy Ba. 2015.
\newblock \href {https://arxiv.org/pdf/1412.6980.pdf} {{ADAM: A Method for
  Stochastic Optimization}}.
\newblock In \emph{3rd International Conference on Learning Representations,
  {ICLR} 2015}, pages 1--15, San Diego, CA, USA.

\bibitem[{Lei and Ruihong(2017)}]{gao2017}
Gao Lei and Huang Ruihong. 2017.
\newblock {Detecting Online Hate Speech Using Context Aware Models}.
\newblock In \emph{Proceedings of the International Conference Recent Advances
  in Natural Language Processing, {RANLP}}, pages 260--266, Varna, Bulgaria.

\bibitem[{Mathew et~al.(2021)Mathew, Saha, Yimam, Biemann, Goyal, and
  Mukherjee}]{Mathew2021}
Binny Mathew, Punyajoy Saha, Seid~Muhie Yimam, Chris Biemann, Pawan Goyal, and
  Animesh Mukherjee. 2021.
\newblock \href {https://ojs.aaai.org/index.php/AAAI/article/view/17745}
  {{HateXplain: A Benchmark Dataset for Explainable Hate Speech Detection}}.
\newblock \emph{Proceedings of the AAAI Conference on Artificial Intelligence},
  35(17):14867--14875.

\bibitem[{Nobata et~al.(2016)Nobata, Tetreault, Thomas, Mehdad, and
  Chang.}]{Nobata2016}
Chikashi Nobata, Joel Tetreault, Achint Thomas, Yashar Mehdad, and Yi~Chang.
  2016.
\newblock {Abusive Language Detection in Online User Content}.
\newblock In \emph{Proceedings of the 25th International Conference on World
  Wide Web.}, pages 145–--153, Montr\'{e}al, QC, Canada.

\bibitem[{Nockleby et~al.(2000)Nockleby, Levy, Karst, and
  editors}]{Nockleby2000}
John~T. Nockleby, Leonard~W. Levy, Kenneth~L. Karst, and Dennis J.~Mahoney
  editors. 2000.
\newblock \emph{Encyclopedia of the American Constitution}.
\newblock Macmillan, 2nd edition.

\bibitem[{Schmidt and Wiegand(2017)}]{schmidt-wiegand-2017-survey}
Anna Schmidt and Michael Wiegand. 2017.
\newblock \href {https://doi.org/10.18653/v1/W17-1101} {{A Survey on Hate
  Speech Detection using Natural Language Processing}}.
\newblock In \emph{Proceedings of the Fifth International Workshop on Natural
  Language Processing for Social Media}, pages 1--10, Valencia, Spain.
  Association for Computational Linguistics.

\bibitem[{Silva et~al.(2021)Silva, Mondal, Correa, Benevenuto, and
  Weber}]{silva2016analyzing}
Leandro Silva, Mainack Mondal, Denzil Correa, Fabrício Benevenuto, and Ingmar
  Weber. 2021.
\newblock \href {https://ojs.aaai.org/index.php/ICWSM/article/view/14811}
  {Analyzing the targets of hate in online social media}.
\newblock In \emph{Proceedings of the International AAAI Conference on Web and
  Social Media}, pages 687--690, Cologne, Germany.

\bibitem[{Williams et~al.(2020)Williams, Burnap, Javed, Liu, and
  Ozalp}]{Williams2020}
Matthew~L. Williams, Pete Burnap, Amir Javed, Han Liu, and Sefa Ozalp. 2020.
\newblock {Hate in the Machine: Anti-Black and Anti-Muslim Social Media Posts
  as Predictors of Offline Racially and Religiously Aggravated Crime}.
\newblock \emph{The British Journal of Criminology}, 60(1):93--117.

\bibitem[{Xiang et~al.(2012)Xiang, Fan, Wang, Hong, and Rose}]{Xiang2012}
Guang Xiang, Bin Fan, Ling Wang, Jason Hong, and Carolyn Rose. 2012.
\newblock {Detecting Offensive Tweets via Topical Feature Discovery over a
  Large Scale Twitter Corpus}.
\newblock In \emph{Proceedings of the 21st ACM international conference on
  Information and knowledge management}, pages 1980–--1984, Maui, HI, USA.

\bibitem[{Xu et~al.(2012)Xu, Jun, Zhu, and Bellmore}]{xu-etal-2012-learning}
Jun-Ming Xu, Kwang-Sung Jun, Xiaojin Zhu, and Amy Bellmore. 2012.
\newblock \href {https://www.aclweb.org/anthology/N12-1084} {{Learning from
  Bullying Traces in Social Media}}.
\newblock In \emph{Proceedings of the 2012 Conference of the North {A}merican
  Chapter of the Association for Computational Linguistics: Human Language
  Technologies}, pages 656--666, Montr{\'e}al, QC, Canada. Association for
  Computational Linguistics.

\bibitem[{Zhong et~al.(2016)Zhong, Li, Squicciarini, Rajtmajer, Griffin,
  Miller, and Caragea}]{Zhong}
Haoti Zhong, Hao Li, Anna Squicciarini, Sarah Rajtmajer, Christopher Griffin,
  David Miller, and Cornelia Caragea. 2016.
\newblock {Content-Driven Detection of Cyberbullying on the Instagram Social
  Network}.
\newblock In \emph{Proceedings of the Twenty-Fifth International Joint
  Conference on Artificial Intelligence (IJCAI-16)}, IJCAI'16, pages
  3952–--3958, New York, NY, USA.

\end{thebibliography}
\bibliographystyle{acl_natbib}

\newpage
\appendix

\section{Appendix}
\label{sec:appendix}
\subsection{Hyperparameter values for experiments}
\label{section:hyperparams}
All the models used the Adam optimizer \cite{kingma2017adam}. Bi-LSTM and BoW used the cross-entropy loss function whereas our BERT models used the sparse categorical and cross-entropy loss function. Further values for the hyperparameters for each experiment are shown in Table \ref{hyperparams}.

\subsubsection{Bi-LSTM}
For all the models except for the model  trained on the Twitter dataset, the architecture consists of an embedding layer followed by two Bi-LSTM layers stacked one after another.
Finally, a Dense layer with a softmax activation function is giving the output class. 

For training with Twitter (both in-domain and domain adaptation), a single Bi-LSTM layer is used.

\subsubsection{BoW}
The BoW model uses two hidden layers consisting of 100 neurons each. 

\subsubsection{BERT}
BERT uses TFBertForSequenceClassification model and BertTokenizer as its tokenizer from the pretrained model bert-base-uncased. 

\begin{table*}
\begin{tabular}{llllll}
\hline
\textbf{Model} & \textbf{Train-Dataset} & \textbf{Test-Dataset} &\textbf{Learning Rate} & \textbf{Epochs} & \textbf{Batch Size}\\
\hhline{|=|=|=|=|=|=|}
BoW & Fox News & Fox News & 1e-03 & 8 & 32 \\
BoW & Twitter & Twitter & 1e-03 & 8 & 32 \\
\hline
BoW & Fox News & Movies & 1e-03 & 8 & 32 \\
BoW & Twitter & Movies & 1e-03 & 8 & 32 \\
\hline
BoW & Movies & Movies & 1e-03 & 8 & 32 \\
\hhline{|=|=|=|=|=|=|}
BERT & Fox News & Fox News & 3e-06  & 17 & 32 \\
BERT & Twitter & Twitter & 3e-06  & 4 & 32 \\
\hline
BERT & Fox News & Movies & 3e-06  & 17 & 32 \\
BERT & Twitter & Movies & 3e-06  & 4 & 32 \\
\hline
BERT & Movies & Movies & 3e-06  & 6 & 32 \\
\hline
BERT & Fox News and Movies & Movies & 3e-06  & 6 & 32 \\
BERT & Twitter and Movies & Movies & 3e-06  & 6 & 32 \\
\hhline{|=|=|=|=|=|=|}
Bi-LSTM & Fox News & Fox News & 1e-03  & 8 & 32 \\
Bi-LSTM & Twitter & Twitter & 1e-03  & 8 & 32 \\
\hline
Bi-LSTM & Fox News & Movies & 1e-03  & 8 & 32 \\
Bi-LSTM & Twitter & Movies & 1e-03  & 8 & 32 \\
\hline
Bi-LSTM & Movies & Movies & 1e-03  & 8 & 32 \\
\hline
\end{tabular}
\caption{
Detailed setups of all applied experiments
}
\label{hyperparams}
\end{table*}

\newpage
\subsection{Additional Performance Metrics for Experiments}
\label{section:detailmetrics}
We report precision, recall, F1-score and macro averaged F1-score for every experiment in Table \ref{classificationReports}.

\begin{table*}
\scalebox{0.9}{
\begin{tabular}{llllllll}
\hline
\textbf{Model} & \textbf{Train-Dataset} & \textbf{Test-Dataset} & \textbf{Category} & \textbf{Precision} & \textbf{Recall} & \textbf{F1-Score} & \textbf{Macro AVG F1}\\
\hhline{|=|=|=|=|=|=|=|=|}
BoW & Fox News & Fox News & normal & 0.81 & 0.84 & 0.83 & 0.63\\
BoW & Fox News & Fox News & hate  & 0.45 & 0.41 & 0.43 & 0.63\\
BoW & Twitter & Twitter & normal  & 0.79 & 0.78 & 0.78 & 0.66\\
BoW & Twitter & Twitter & offensive  & 0.90 & 0.95 & 0.93 & 0.66\\
BoW & Twitter & Twitter & hate  & 0.43 & 0.18 & 0.26 & 0.66\\
\hline
BoW & Fox News & Movies & normal & 0.84 & 0.87 & 0.86 & 0.51\\
BoW & Fox News & Movies & hate  & 0.16 & 0.13 & 0.15 & 0.51\\
BoW & Twitter & Movies & normal  & 0.96 & 0.46 & 0.62 & 0.37\\
BoW & Twitter & Movies & offensive  & 0.20 & 0.82 & 0.32 & 0.37\\
BoW & Twitter & Movies & hate  & 0.11 & 0.24 & 0.15 & 0.37\\
\hline
BoW & Movies & Movies & normal  & 0.93 & 0.97 & 0.95 & 0.64\\
BoW & Movies & Movies & offensive  & 0.65 & 0.56 & 0.59 & 0.64\\
BoW & Movies & Movies & hate  & 0.56 & 0.28 & 0.37 & 0.64\\
\hline
BERT & Fox News & Fox News & normal & 0.84 & 0.87 & 0.86 & 0.68\\
BERT & Fox News & Fox News & hate  & 0.57 & 0.46 & 0.51 & 0.68\\
BERT & Twitter & Twitter & normal  & 0.88 & 0.91 & 0.89 & 0.76\\
BERT & Twitter & Twitter & offensive  & 0.94 & 0.97 & 0.95 & 0.76\\
BERT & Twitter & Twitter & hate  & 0.59 & 0.34 & 0.43 & 0.76\\
\hline
BERT & Fox News & Movies & normal & 0.88 & 0.90 & 0.89 & 0.64\\
BERT & Fox News & Movies & hate  & 0.40 & 0.37 & 0.39 & 0.64\\
BERT & Twitter & Movies & normal  & 0.98 & 0.92 & 0.95 & 0.77\\
BERT & Twitter & Movies & offensive  & 0.63 & 0.90 & 0.74 & 0.77\\
BERT & Twitter & Movies & hate  & 0.63 & 0.63 & 0.63 & 0.77\\
\hline
BERT & Movies & Movies & normal & 0.97 & 0.98 & 0.97 & 0.81\\
BERT & Movies & Movies & offensive  & 0.80 & 0.76 & 0.78 & 0.81\\
BERT & Movies & Movies & hate  & 0.79 & 0.68 & 0.68 & 0.81\\
\hline
BERT & Fox News and Movies & Movies & normal & 0.97 & 0.97 & 0.97 & 0.89\\
BERT & Fox News and Movies & Movies & hate  & 0.83 & 0.81 & 0.82 & 0.89\\
BERT & Twitter and Movies & Movies & normal  & 0.97 & 0.97 & 0.97 & 0.77\\
BERT & Twitter and Movies & Movies & offensive  & 0.76 & 0.76 & 0.75 & 0.77\\
BERT & Twitter and Movies & Movies & hate  & 0.57 & 0.73 & 0.59 & 0.77\\
\hline
Bi-LSTM & Fox News & Fox News & normal & 0.83 & 0.72 & 0.77 & 0.62\\
Bi-LSTM & Fox News & Fox News & hate  & 0.39 & 0.55 & 0.46 & 0.62\\
Bi-LSTM & Twitter & Twitter & normal  & 0.74 & 0.78 & 0.76 & 0.66\\
Bi-LSTM & Twitter & Twitter & offensive  & 0.91 & 0.91 & 0.91 & 0.66\\
Bi-LSTM & Twitter & Twitter & hate  & 0.31 & 0.31 & 0.31 & 0.66\\
\hline
Bi-LSTM & Fox News & Movies & normal & 0.85 & 0.81 & 0.83 & 0.51\\
Bi-LSTM & Fox News & Movies & hate  & 0.17 & 0.20 & 0.18 & 0.51\\
Bi-LSTM & Twitter & Movies & normal  & 0.96 & 0.50 & 0.66 & 0.38\\
Bi-LSTM & Twitter & Movies & offensive  & 0.22 & 0.79 & 0.34 & 0.38\\
Bi-LSTM & Twitter & Movies & hate  & 0.10 & 0.33 & 0.16 & 0.38\\
\hline
Bi-LSTM & Movies & Movies & normal  & 0.94 & 0.97 & 0.95 & 0.71\\
Bi-LSTM & Movies & Movies & offensive  & 0.67 & 0.60 & 0.63 & 0.71\\
Bi-LSTM & Movies & Movies & hate  & 0.73 & 0.49 & 0.56 & 0.71\\
\hline
HateXplain & - & Movies & normal & 0.88 & 0.98 & 0.93 & 0.66\\
HateXplain & - & Movies & offensive  & 0.62 & 0.17 & 0.27 & 0.66\\
HateXplain & - & Movies & hate  & 0.89 & 0.68 & 0.77 & 0.66\\
\hline
\end{tabular}
}
\caption{
Detailed results of all applied experiments
}
\label{classificationReports}
\end{table*}

\end{document}